# Robust Hyperspectral Unmixing with Correntropy based Metric

Ying Wang, Chunhong Pan, Shiming Xiang, and Feiyun Zhu

## Abstract

Hyperspectral unmixing is one of the crucial steps for many hyperspectral applications. The problem of hyperspectral unmixing has proven to be a difficult task in unsupervised work settings where the endmembers and abundances are both unknown. What is more, this task becomes more challenging in the case that the spectral bands are degraded with noise. This paper presents a robust model for unsupervised hyperspectral unmixing. Specifically, our model is developed with the correntropy based metric where the non-negative constraints on both endmembers and abundances are imposed to keep physical significance. In addition, a sparsity prior is explicitly formulated to constrain the distribution of the abundances of each endmember. To solve our model, a half-quadratic optimization technique is developed to convert the original complex optimization problem into an iteratively re-weighted NMF with sparsity constraints. As a result, the optimization of our model can adaptively assign small weights to noisy bands and give more emphasis on noise-free bands. In addition, with sparsity constraints, our model can naturally generate sparse abundances. Experiments on synthetic and real data demonstrate the effectiveness of our model in comparison to the related state-of-the-art unmixing models.

Ying Wang, Chunhong Pan, Shiming Xiang and Feiyun Zhu are with the National Laboratory of Pattern Recognition, Institute of Automation, Chinese Academy of Sciences (e-mail: {ywang, chpan, smxiang, fyzhu}@nlpr.ia.ac.cn).

# I. Introduction

Hyperspectral remote sensing has wide applications since it provides digital images in many contiguous and very narrow spectral bands. However, due to *the low spatial resolution of hyperspectral sensor, microscopic material mixing*, and *multiple scattering*, spectra measured by hyperspectral sensor are likely to be mixed by several materials [1], [2], [3]. Accordingly, mixture spectra present a challenge to hyperspectral image analysis because their spectral signatures do not correspond to any single well-defined material [1]. Thus, hyperspectral unmixing is a significant step for hyperspectral image analysis. The task of hyperspectral unmixing is to separate the mixed pixel spectra from a hyperspectral image into a collection of constituent spectra, called *endmembers* and a set of corresponding fractions, called *abundances* [1].

In general, there are two kinds of spectral unmixing model: linear and nonlinear. Linear mixing model (LMM) is the most popular model for hyperspectral unmixing. It assumes that the spectrum of a given pixel is a linear combination of the endmembers [1], [3]. Due to its practical advantages, there are a lot of unmixing models proposed based on LMM [4], [5], [6], [7], [8], [9], [10], [11]. Conversely, nonlinear unmixing model [3], [12], [13] describes mixed spectra by assuming that the observed pixel is generated from a nonlinear function of the abundance associated with the endmembers.

In literature, conventional algorithms for hyperspectral unmixing involve two steps: *endmember extraction* and *mixed-pixel decomposition*. For endmember extraction, a majority of algorithms have been designed under the pure pixel assumption [14], [15], [16], [17], [18]. N-FINDR [14] and Vertex component analysis (VCA) [17] are two typical this kind of algorithms. N-FINDR [14] aims at looking for the pure pixels in the hyperspectral data that determine the maximum volume. VCA [17] considers that the endmembers are the vertices of a simplex and extracts the endmembers iteratively by projecting the data onto the direction which is orthogonal to the subspace spanned by the already extracted endmembers. All these algorithms assume the presence of pure pixels for each endmember in the original hyperspectral data. Once the endmembers have been identified, the corresponding abundances can be estimated by solving a constrained least square problem. The constraints include the non-negativity constraint, the sum-to-one constraint, and so on [4], [19].

Since the assumption of pure pixel presence is usually unreliable, a number of blind source separation algorithms have been proposed to obtain the endmembers and the abundances simultaneously [20], [5], [21], [9], [11]. Among these algorithms, non-negative matrix factroization (NMF) shows significant potential for hyperspectral unmixing [8], [5], [11], [6]. NMF decomposes a non-negative hyperspectral data matrix into the product of two non-negative matrices, which are the endmembers and the abundances respectively. NMF has two main advantages for hyperspectral unmixing. First, the non-negative constraint on both endmembers and abundances is physically consistent with the mixing pixels in real world. Second, NMF usually provides a parts-based representation of the data, making the decomposition matrices more intuitive and interpretable [22], [23]. There are several extension models based on NMF proposed in hyperspectral unmixing literature [8], [5], [9], [11], [7]. Specifically, in [5], the authors introduce the piecewise smoothness of spectral data and sparseness of abundance fraction to NMF. In [7], the authors extend the NMF method by incorporating the $\ell_{1/2}$ sparsity constraint. These algorithms, however, take no account of the noise in each band in the hyerspectral data. Actually, in hyperspectral images, due to the distinctive properties of the hyperspectral sensor in each wavelengths, the noise intensity in different bands is different. Thus, these algorithms may fail to obtain the accurate endmembers and abundances.

Practically, due to the variations in atmospheric conditions and the properties of the hyperspectral sensor in each wavelengths, the noisy bands commonly exist in hyperspectral data [24]. Furthermore, the noise level differs from band to band (as can be seen from Fig. 4). Naturally, the intuitive motivation of this paper is that the bands which have high-noise-intensity should have small contributions for hyperspectral unmixing. This observation motivates us to propose a robust unmixing algorithm based on correntropy based metric [25], [26] to take into account the noise of each band. It can adaptively and automatically assign small weights to noisy bands. Thus, the proposed algorithm is much more insensitive to noisy bands.

The main contributions of this paper are highlighted as follows:

1) A robust hyperspectral unmixing model is proposed based on the correntropy along with

imposing non-negative constraints and the use of a sparse penalty. Beyond directly using the traditional square of the Euclidean distance, the objective function of the proposed model is developed based on correntropy based metric, which can adaptively assign small weights to noisy bands and give more emphasis on noise-free bands. As a result, our model can effectively cope with the noisy bands in hyperspectral data. Experiments on synthetic and real data illustrate the effectiveness of our model.

2) An optimization algorithm of our model is proposed based on half-quadratic optimization technique. It converts the original nonlinear and nonconvex optimization problem into an iteratively re-weighted NMF with sparsity constraints, which can be easily solved by multiplicative update algorithm. Besides, the convergence of this optimization algorithm is proven theoretically.

The remainder of this paper is structured as follows. We first give a brief introduction to the LMM and its extensions in Section II. In Section III, we propose our robust unmixing model based on correntropy based metric and develop a convergent algorithm for its optimization. Section IV reports the experimental results and performs algorithmic comparisons with other related state-of-the-art techniques. Conclusions are drawn in Section V.

## II. RELATED WORKS

### A. Linear Mixture Model

Due to its physical effectiveness and mathematical simplicity [2], [3], Linear Mixture Model (LMM) is commonly used for hyperspectral unmixing. LMM assumes that the observed pixel is approximated by a linear combination of a small number of endmembers. The mathematical formulation of the LMM is given by

$$\mathbf{y} = \sum_{k=1}^{K} w_k \mathbf{x}_k + \mathbf{e}, \tag{1}$$

where $\mathbf{y} \in \mathbb{R}^D$ denotes an observed pixel, $\mathbf{x}_k \in \mathbb{R}^D$ is the $k$-th endmember, $w_k$ is the abundance fraction for the $k$-th endmember and $\mathbf{e} \in \mathbb{R}^D$ is an additive noise.

Considering a hyperspectral image of $N$ pixels, its matrix notation is $\mathbf{Y} = [\mathbf{y}_1, \cdots, \mathbf{y}_N] \in \mathbb{R}^{D \times N}$, then LMM can be rewritten as

$$\mathbf{Y} = \mathbf{XW} + \mathbf{E}, \tag{2}$$

where $\mathbf{X} = [\mathbf{x}_1, \cdots, \mathbf{x}_K] \in \mathbb{R}^{D \times K}$ is the endmember matrix, $\mathbf{W} = [\mathbf{w}_1, \cdots, \mathbf{w}_N] \in \mathbb{R}^{K \times N}$ is the abundance matrix, and $\mathbf{E} \in \mathbb{R}^{D \times N}$ represents the additive noise. The number of endmembers $K$ is often much smaller than the number of band $N$. As mentioned in [1], for spectral unmixing problem, LMM should satisfy two physical constraints, referred to as the non-negative constraint and the abundance sum-to-one constraint, which can be formalized as follows:

$$\mathbf{X}_{dk} \geqslant 0, \ \ \mathbf{W}_{kn} \geqslant 0, \ \ \sum_{k=1}^{K} \mathbf{W}_{kn} = 1.$$

*B. Non-negative Matrix Factorization*

Recently, NMF has received much attention [8], [5], [11], [6] for hyperspectral unmixing because of its non-negative nature. Essentially, NMF is a LMM model with non-negative constraints. It aims at decomposing the given non-negative matrix $\mathbf{Y}$ into two non-negative matrices $\mathbf{X}$ and $\mathbf{W}$. The objective function of NMF is defined as follows:

$$\min_{\mathbf{X},\mathbf{W}} \|\mathbf{Y} - \mathbf{XW}\|_F^2, \ \ \ s.t. \ \mathbf{X}_{dk} \geqslant 0, \mathbf{W}_{kn} \geqslant 0, \tag{3}$$

where$\| \cdot \|_F$ denotes the frobenius norm.

There are numerous optimization algorithms to solve the above problem (3). The most famous one is the multiplicative update algorithm which was proposed by Lee and Seung [22]. It has been proved that the multiplicative update algorithm can find local minima of the objective function (3). There are also several other optimization algorithms, such as the alternation non-negative least least squares [27], the projected gradient descent [28] and the active-set method [29], proposed to solve the NMF problem.

The non-negative constrains on $\mathbf{X}$ and $\mathbf{W}$ of NMF only allow additive combinations among different bases to represent the sample, which makes NMF learn a parts-based representation of data [22], [23]. The advantage of this parts-based representation has been successfully applied in many real-world problems such as document clustering [30], [31], hyperspectral unmixing [32], [5], [7], [11] and face recognition [33], [34].

However, due to the nonconvexity of the objective function of NMF, there are a lot of local minima of the objective function (3). In order to reduce the feasible solution set, several constraints are introduced to NMF framework [35], [32]. In summary, these extensions can be expressed as the following objective function

$$\min_{\mathbf{X},\mathbf{W}} \|\mathbf{Y} - \mathbf{X}\mathbf{W}\|_F^2 + \lambda J_{\mathbf{X}}(\mathbf{X}) + \beta J_{\mathbf{W}}(\mathbf{W}),$$
$$s.t. \ \mathbf{X}_{dk} \geqslant 0, \mathbf{W}_{kn} \geqslant 0, \tag{4}$$

where $J_{\mathbf{X}}(\mathbf{X})$ and $J_{\mathbf{W}}(\mathbf{W})$ denote the regularization terms for $\mathbf{X}$ and $\mathbf{W}$, respectively, and $\lambda \geqslant 0, \beta \geqslant 0$ are the regularization parameters.

Due to the physical consistency with unmixing problem, NMF and its extension models have been widely applied to hyperspectral unmixing. However, most of these methods do not take into account the difference of noise intensity in each band. These algorithms consider that all of the bands in the hypersepctral data have the same signal-to-noise ratio. Actually, the noise intensity in different bands is different in hyperspectral data. Accordingly, along with the line of correntropy based metric, in this paper we propose a robust unmixing method to suppress the contributions of the bands with high-noise-intensity.

## III. Robust unmixing based on correntropy based metric

In this section, we first describe the correntropy. Then, we propose our robust unmixing model based on correntropy based metric. Finally, we present the optimization algorithm of our model and summarize its implementation details.

### A. Correntropy

In hyperspectral images, the noise levels of different bands are usually different (as can be seen from Fig. 4). Meanwhile, noise is usually difficult for computer to predict in advance. These noisy bands will severely affect the unmixing results. Recently, in information theoretical learning [36], correntropy was proposed to obtain robust analysis [37]. It has shown many advantages in robust learning and pattern recognition [26], [38]. Correntropy is a nonlinear similarity measure between two random variables $\mathbf{x}$ and $\mathbf{y}$, which is defined by

$$C_\sigma(\mathbf{x}, \mathbf{y}) = \mathbb{E}\left[\kappa_\sigma(\mathbf{x} - \mathbf{y})\right],$$  (5)

where $\kappa_\sigma(\cdot)$ denotes a kernel function that satisfies Mercer's Theorem [39] and $\mathbb{E}$ is the expectation operator.

Actually, the joint probability density function of $\mathbf{x}$ and $\mathbf{y}$ is practically unknown. Only a finite number of data $\{(\mathbf{x}_n, \mathbf{y}_n)\}_{n=1}^N$ are avaiable. Thus, the sample estimator of correntropy is estimate by

$$\hat{C}_\sigma(\mathbf{x}, \mathbf{y}) = \frac{1}{N} \sum_{n=1}^N \kappa_\sigma(\mathbf{x}_n - \mathbf{y}_n).$$  (6)

In this work, we adopt the Gaussian kernel function $\kappa_\sigma(\mathbf{x} - \mathbf{y}) = \exp\left(-\frac{\|\mathbf{x} - \mathbf{y}\|_2^2}{\sigma^2}\right)$.

Actually, correntropy has a close relationship with Welsch M-estimator [40]. In [25], the authors have proved that correntropy is a robust function for linear and nonlinear regression.

### B. Proposed model

For convenience, a hyperspectral image with $N$ pixels and $D$ bands is denoted as $\mathbf{Y} = [\mathbf{y}_1, \cdots, \mathbf{y}_N] = \left[\mathbf{y}^1, \cdots, \mathbf{y}^D\right]^T \in \mathbb{R}^{D \times N}$, where the pixel $\mathbf{y}_n \in \mathbb{R}^D$ is the $n$-th column of $\mathbf{Y}$ and the band $\mathbf{y}^d \in \mathbb{R}^N$ is represented by the $d$-th row of $\mathbf{Y}$. Suppose the observed hyperspectral image $\mathbf{Y}$ is linearly mixed by the endmembers $\mathbf{X} = [\mathbf{x}_1, \cdots, \mathbf{x}_K] \in \mathbb{R}^{D \times K}$ and the abundance $\mathbf{W} = [\mathbf{w}_1, \cdots, \mathbf{w}_N] \in \mathbb{R}^{K \times N}$. The commonly used cost function for LMM is the square of the Euclidean distance between $\mathbf{Y}$ and $\mathbf{XW}$, which is defined as

$$L(\mathbf{X}, \mathbf{W}) = \|\mathbf{Y} - \mathbf{XW}\|_F^2.$$  (7)

However, the solution based on square of the Euclidean distance has been recognized to be sensitive to noise and outliers [41]. The existence of noise in a hyperspectral image will limit the precision of spectral unmixing. Since hyperspectral images have hundreds of bands and the noise level in different bands is different, we define the following corrtentropy based model.

$$
\begin{aligned}
L(\mathbf{X}, \mathbf{W}) &= \sum_{d=1}^{D} \left\{ -\kappa_{\sigma} \left( \mathbf{y}^d - (\mathbf{XW})^d \right) \right\} \\
&= \sum_{d=1}^{D} \left\{ -\exp \left( -\frac{\left\| \mathbf{y}^d - (\mathbf{XW})^d \right\|_2^2}{\sigma^2} \right) \right\},
\end{aligned}
\tag{8}
$$

where $(\mathbf{XW})^d$ denotes the $d$-th row of matrix $\mathbf{XW}$. Different from the NMF based methods where all bands will contribute equally to the objective function, our model treats individual band differently and gives more emphasis on those bands that have more exactly linear reconstruction. If the $d$-th band $\mathbf{y}^d$ is degraded by serious noise, then this band will yield a large reconstruction error. According to the properties of Welsch M-estimator [40], the $d$-th band will provide small contribution to the objective function in Eq. (8). This implies that if there are noise-corrupted bands, those bands will have small contributions to the objective function.

It is worth to point out that the above model defined in (8) is different from the traditional robust estimation methods [41], which consider that there are outliers in the samples (pixels). In practical situations, there are usually noisy bands in the hyperspectral images. Consequently, our model considers that there are "outlier" bands, which is more physically consistent with the real situation in hyperspectral data.

In addition, we take into account the sparseness constraint of the abundance. In most cases, the abundance distribution of any endmember does not apply to the whole scene. This implies that the observed pixel is usually mixed by only a few endmembers. Therefore, the abundance fractions of each pixel have a sparsity prior. In order to enforce sparsity, an $\ell_0$-norm penalty term $\sum_{n=1}^{N} \|\mathbf{w}_n\|_0$ can be incorporated into our model. However, the $\ell_0$ minimization problem is NP hard, which means that it is very hard to solve.

Recent development in the theory of sparse representation and compressed sensing [42], [43], [44] shows that if the solution is sparse enough, then the solution of $\ell_0$ minimization problem is equal to the solution of the $\ell_1$ problem. Besides, since its convexity property, $\ell_1$ minimization problem is much easier to be solved. Thus, we introduce the following $\ell_1$-norm penalty term into our model.

$$\sum_{n=1}^{N} \|\mathbf{w}_n\|_1 . \tag{9}$$

Combining the correntropy based loss term (8) and the $\ell_1$ sparse penalty term (9), the cost function of our robust unmixing model is defined as follows:

$$\mathcal{G}(\mathbf{X}, \mathbf{W}) = \sum_{d=1}^{D} \left\{ -\exp\left( -\frac{\left\| \mathbf{y}^d - (\mathbf{X}\mathbf{W})^d \right\|_2^2}{\sigma^2} \right) \right\} + 2\lambda \sum_{n=1}^{N} \|\mathbf{w}_n\|_1 , \tag{10}$$

where $\lambda \geqslant 0$ is the regularization parameter. It is worth noting that the minimization problem (10) is subjected to $\mathbf{X}_{dk} \geqslant 0$ and $\mathbf{W}_{kn} \geqslant 0$, which are the non-negative constraints. Here, $\mathbf{X}_{dk}$ denotes the spectrum of the $k$-th endmember in $d$-th band and $\mathbf{W}_{kn}$ represents the abundance fraction of the $n$-th pixel corresponding to $k$-th endmember. It is obvious that our model is essentially a correntropy based NMF (CENMF) along with an $\ell_1$ sparse penalty term. We call this model $\ell_1$-CENMF.

### C. Optimization

Obviously the objective function in (10) is nonconvex and nonlinear with respect to $\mathbf{X}$ and $\mathbf{W}$ together, and it is difficult to optimize directly. Here, we propose an alternative iterative algorithm for its optimization.

Our algorithm utilizes the half-quadratic technique [45], [46] to solve the optimization problem (10). According to the property of the convex conjugated function, we have [47]:

**Proposition 1.** *There exists a convex conjugated function* $\psi(\cdot)$ *of* $g(x) = \exp(-x)$*, such that*

$$g(x) = \max_u \left( ux - \psi(u) \right), \tag{11}$$

*and for a fixed* $x$*, the maximum is reached at* $u = -g(x)$*.*

According to the proposition 1, substituting $x$ with $\frac{\|\mathbf{x}\|_2^2}{\sigma^2}$, we get the following proposition

**Proposition 2.** *There exists a convex conjugated function $\psi(\cdot)$ of $\kappa_\sigma(\mathbf{x}) = \exp(-\frac{\|\mathbf{x}\|_2^2}{\sigma^2})$ such that*

$$-\kappa_\sigma(\mathbf{x}) = \min_u \left( u \frac{\|\mathbf{x}\|_2^2}{\sigma^2} + \psi(-u) \right), \tag{12}$$

*and for a fixed $\mathbf{x}$, the minimum is reached at $u = \kappa_\sigma(\mathbf{x})$[1].*

Substituting (12) into (10), we have the following augmented objective function in an enlarged parameter space

$$\widehat{\mathcal{G}}(\mathbf{X}, \mathbf{W}, \mathbf{u}) = \sum_{d=1}^{D} \left( u_d \frac{\left\| \mathbf{y}^d - (\mathbf{X}\mathbf{W})^d \right\|_2^2}{\sigma^2} + \psi(-u_d) \right) + 2\lambda \sum_{n=1}^{N} \|\mathbf{w}_n\|_1, \tag{13}$$

where $\mathbf{u} = [u_1, u_2, \cdots, u_D]^T$ is the auxiliary vector introduced by half-quadratic optimization. According to the proposition 2, for a fixed $\mathbf{X}, \mathbf{W}$, we obtain the following equation,

$$\mathcal{G}(\mathbf{X}, \mathbf{W}) = \min_{\mathbf{u}} \widehat{\mathcal{G}}(\mathbf{X}, \mathbf{W}, \mathbf{u}). \tag{14}$$

Note that the minimization is reached at $u_d = \exp\left(-\frac{\left\| \mathbf{y}^d - (\mathbf{X}\mathbf{W})^d \right\|_2^2}{\sigma^2}\right)$.

Then, we can conclude that minimizing $\mathcal{G}(\mathbf{X}, \mathbf{W})$ is identical to minimizing the augmented function $\widehat{\mathcal{G}}(\mathbf{X}, \mathbf{W}, \mathbf{u})$. The augmented function $\widehat{\mathcal{G}}(\mathbf{X}, \mathbf{W}, \mathbf{u})$ can be optimized by the following alternate minimization:

1) Fixed $\mathbf{W}$ and $\mathbf{X}$, minimizing $\widehat{\mathcal{G}}(\mathbf{X}, \mathbf{W}, \mathbf{u})$ with respect to $\mathbf{u}$, which obtains

$$u_d = \exp\left(-\frac{\left\| \mathbf{y}^d - (\mathbf{X}\mathbf{W})^d \right\|_2^2}{\sigma^2}\right). \tag{15}$$

This conclusion can be easily obtained according to proposition 2.

---

[1]Here, we replace $u = -u$ in Eq. (12).

2) Fixed $\mathbf{u}$, minimizing $\widehat{\mathcal{G}}(\mathbf{X}, \mathbf{W}, \mathbf{u})$ with respect to $\mathbf{W}$ and $\mathbf{X}$, which is equivalent to

$$\min_{\mathbf{W}, \mathbf{X}} \left\{ \mathrm{Tr} \left( (\mathbf{Y} - \mathbf{XW})^T \mathbf{U} (\mathbf{Y} - \mathbf{XW}) \right) + 2\lambda \sum_{n=1}^{N} \|\mathbf{w}_n\|_1 \right\},$$

$$s.t. \ \mathbf{X}_{dk} \geqslant 0, \mathbf{W}_{kn} \geqslant 0, \tag{16}$$

where $\mathrm{Tr} \left( \cdot \right)$ denotes the trace of a matrix and $\mathbf{U}$ is a diagonal matrix whose diagonal element is $\mathbf{U}_{dd} = \frac{u_d}{\sigma^2}$. It is clear that the optimization problem in (16) is a weighted NMF with an $\ell_1$ sparse constraint.

It is easy to prove that the objective function $\mathcal{G}$ in (10) is nonincreasing under the above iterative procedure.

*Proof:* Let $\mathbf{W}^{(t)}$, $\mathbf{X}^{(t)}$ and $\mathbf{u}^{(t)}$ denote the solutions at $t$-th iteration. According to Eq. (14), it is obvious that

$$\mathcal{G} \left( \mathbf{X}^{(t)}, \mathbf{W}^{(t)} \right) = \widehat{\mathcal{G}} \left( \mathbf{X}^{(t)}, \mathbf{W}^{(t)}, \mathbf{u}^{(t)} \right). \tag{17}$$

According to the step 1 of the above alternate iteration minimization, because $\mathbf{u}^{(t+1)}$ is the solution of minimizing $\hat{\mathcal{G}} \left( \mathbf{X}^{(t)}, \mathbf{W}^{(t)}, \mathbf{u} \right)$, then we obtain

$$\widehat{\mathcal{G}} \left( \mathbf{X}^{(t)}, \mathbf{W}^{(t)}, \mathbf{u}^{(t+1)} \right) \leqslant \widehat{\mathcal{G}} \left( \mathbf{X}^{(t)}, \mathbf{W}^{(t)}, \mathbf{u}^{(t)} \right). \tag{18}$$

Similarly, according the step 2, because $\mathbf{X}^{(t+1)}$ and $\mathbf{W}^{(t+1)}$ are the solutions of minimizing $\widehat{\mathcal{G}} \left( \mathbf{X}, \mathbf{W}, \mathbf{u}^{(t+1)} \right)$, we have

$$\widehat{\mathcal{G}} \left( \mathbf{X}^{(t+1)}, \mathbf{W}^{(t+1)}, \mathbf{u}^{(t+1)} \right) \leqslant \widehat{\mathcal{G}} \left( \mathbf{X}^{(t)}, \mathbf{W}^{(t)}, \mathbf{u}^{(t+1)} \right). \tag{19}$$

Summarizing Eqs. (17), (18) and (19), we have

$$\mathcal{G} \left( \mathbf{X}^{(t+1)}, \mathbf{W}^{(t+1)} \right) \leqslant \mathcal{G} \left( \mathbf{X}^{(t)}, \mathbf{W}^{(t)} \right). \tag{20}$$

In this way, the objective function $\mathcal{G}$ in Eq. (10) is non-increasing under the above alternative iterative procedure. ∎

*D. Implementation details*

Note that the key task of minimizing the objective function $\mathcal{G}$ in Eq. (10) is to solve the problem (16) in the step 2 of the above alternative iterative procedure. It is clear that this optimization problem (16) is a weighted $\ell_1$-NMF, which can be further formulated as follows:

$$\min_{\mathbf{W}, \widehat{\mathbf{X}}} \left\{ \mathrm{Tr} \left( \left( \widehat{\mathbf{Y}} - \widehat{\mathbf{X}} \mathbf{W} \right)^T \left( \widehat{\mathbf{Y}} - \widehat{\mathbf{X}} \mathbf{W} \right) \right) + 2\lambda \sum_{n=1}^{N} \| \mathbf{w}_n \|_1 \right\},$$

$$s.t. \ \widehat{\mathbf{X}}_{dk} \geqslant 0, \mathbf{W}_{kn} \geqslant 0, \tag{21}$$

where $\widehat{\mathbf{Y}} = \mathbf{U}^{\frac{1}{2}} \mathbf{Y}$ and $\widehat{\mathbf{X}} = \mathbf{U}^{\frac{1}{2}} \mathbf{X}$.

The above problem (21) is a typical $\ell_1$ sparse constrained NMF, which can be solved easily by multiplicative update algorithm [48].

Let $\Phi_{dk}$ and $\Psi_{kn}$ be the Lagrange multiplier for constraint $\widehat{\mathbf{X}}_{dk} \geqslant 0$ and $\mathbf{W}_{kn} \geqslant 0$ respectively, and $\Phi = [\Phi_{dk}] \in \mathbb{R}^{D \times K}$, $\Psi = [\Psi_{kn}] \in \mathbb{R}^{K \times N}$. Then the Lagrange is

$$\mathcal{L} = \mathrm{Tr} \left( \widehat{\mathbf{Y}} \widehat{\mathbf{Y}}^T \right) - 2\mathrm{Tr} \left( \widehat{\mathbf{X}} \mathbf{W} \widehat{\mathbf{Y}}^T \right)$$

$$+ \mathrm{Tr} \left( \widehat{\mathbf{X}} \mathbf{W} \mathbf{W}^T \widehat{\mathbf{X}}^T \right) + 2\lambda \sum_{d=1}^{D} \| \mathbf{w}_d \|_1$$

$$+ \mathrm{Tr} \left( \Phi \widehat{\mathbf{X}}_{dk} \right) + \mathrm{Tr} \left( \Psi \mathbf{W} \right). \tag{22}$$

Then, the partial derivatives of $\mathcal{L}$ with respect to $\widehat{\mathbf{X}}$ and $\mathbf{W}$ are

$$\frac{\partial \mathcal{L}}{\partial \widehat{\mathbf{X}}} = -2 \widehat{\mathbf{Y}} \mathbf{W}^T + 2 \widehat{\mathbf{X}} \mathbf{W} \mathbf{W}^T + \Phi, \tag{23}$$

$$\frac{\partial \mathcal{L}}{\partial \mathbf{W}} = -2 \widehat{\mathbf{X}}^T \widehat{\mathbf{Y}} + 2 \widehat{\mathbf{X}}^T \widehat{\mathbf{X}} \mathbf{W} + 2\lambda + \Psi. \tag{24}$$

Using the Karush-Kuhn-Tucker conditions $\Phi_{dk} \widehat{\mathbf{X}}_{dk} = 0$ and $\Psi_{kn} \mathbf{W}_{kn} = 0$, we can obtain the following equations

$$- \left( \widehat{\mathbf{Y}} \mathbf{W}^T \right)_{dk} \widehat{\mathbf{X}}_{dk} + \left( \widehat{\mathbf{X}} \mathbf{W} \mathbf{W}^T \right)_{dk} \widehat{\mathbf{X}}_{dk} = 0, \tag{25}$$

$$- \left( \widehat{\mathbf{X}}^T \widehat{\mathbf{Y}} \right)_{kn} \mathbf{W}_{kn} + \left( \widehat{\mathbf{X}}^T \widehat{\mathbf{X}} \mathbf{W} \right)_{kn} \mathbf{W}_{kn} + \lambda \mathbf{W}_{kn} = 0. \tag{26}$$

Thus, the optimization problem (21) can be solved by the following multiplicative update rules

$$\widehat{\mathbf{X}}_{dk} = \widehat{\mathbf{X}}_{dk} \frac{\left(\widehat{\mathbf{Y}}\mathbf{W}^T\right)_{dk}}{\left(\widehat{\mathbf{X}}\mathbf{W}\mathbf{W}^T\right)_{dk}}, \tag{27}$$

$$\mathbf{W}_{kn} = \mathbf{W}_{kn} \frac{\left(\widehat{\mathbf{X}}^T\widehat{\mathbf{Y}}\right)_{kn}}{\left(\widehat{\mathbf{X}}^T\widehat{\mathbf{X}}\mathbf{W}\right)_{kn} + \lambda}. \tag{28}$$

The detail steps of optimizing the problem (21) are presented in Algorithm 1.

---

**Algorithm 1** Algorithm for optimizing the problem (21).

---

**Input:** Input data matrix $\widehat{\mathbf{Y}}$, the initial $\widehat{\mathbf{X}}$ and $\mathbf{W}$.
**Output:** Two matrices $\widehat{\mathbf{X}}$ and $\mathbf{W}$.
 1: **repeat**
 2:     update $\widehat{\mathbf{X}}$ by Eq. (27).
 3:     update $\mathbf{W}$ by Eq. (28).
 4: **until** convergence

---

In summary, the optimization procedure of our model is outlined in Algorithm 2. As mentioned previously, the optimization procedure of our model is an adaptive weighted $\ell_1$-NMF problem, which can automatically assign small weights to noisy bands. Specifically, the entries of the auxiliary vector $\mathbf{u} = [u_1, \cdots, u_D]^T$ are viewed as the weights. Here, $u_d$ is the weight of the $d$-th band. If the $d$-th band of a hyperspectral image contains serious noise, then this band will yield a large reconstruction error to the LMM model. Consequently, the $d$-th band will be assigned a small weight according to Eq. (15).

Finally, we analyze the computation complexity of the proposed algorithm. It is clear that the computation complexity of optimizing the $\ell_1$-NMF problem (21) in Algorithm 1 is $\mathcal{O}(DNK)$. According to the description in Section III-B, the optimization procedure of our model is equivalent to the iteratively re-weighted NMF with an $\ell_1$ sparse constraint. This implies that if the iteration of Algorithm 2 ends after $T$ steps, the complexity of our $\ell_1$-CENMF is $\mathcal{O}(TDNK)$.

---

**Algorithm 2** Algorithm of our $\ell_1$-CENMF unmixing method.

---

**Input:** Hyperspectral data $\mathbf{Y} \in \mathbb{R}_+^{D \times N}$, the number of endmembers $K$, the sparse regularization parameter $\lambda$ and the stopping criteria $\epsilon$.

**Output:** The endmember matrix $\mathbf{X} \in \mathbb{R}_+^{D \times K}$ and the abundance matrix $\mathbf{W} \in \mathbb{R}_+^{K \times N}$.

  1: **Initialization**. Initialize $\mathbf{X}$ with VCA [17], and then initialize $\mathbf{W}$ with the method in [49]. Initialize $u_d = 1, d = \{1, \cdots, D\}$.

  2: **while** $\|\mathcal{G}^{(t+1)} - \mathcal{G}^{(t)}\| \geqslant \epsilon$ **do**

  3:      $\{\mathcal{G}^{(t)}$ denotes the value of function $\mathcal{G}(\mathbf{X}, \mathbf{W})$ defined in Eq. (10) at the $t$-th iteration.$\}$

  4:      compute $\widehat{\mathbf{Y}} = \mathbf{U}^{\frac{1}{2}}\mathbf{Y}, \widehat{\mathbf{X}} = \mathbf{U}^{\frac{1}{2}}\mathbf{X}$.

  5:      solve the optimization problem (21) according to Algorithm 1, and get the solution $\widehat{\mathbf{X}}$ and $\mathbf{W}$.

  6:      set $\mathbf{X} = \mathbf{U}^{-\frac{1}{2}}\widehat{\mathbf{X}}$.

  7:      update vector $\mathbf{u}$ and Gaussian kernel parameter $\sigma$ according to (15) and (29) respectively.

  8: **end while**

---

## IV. EXPERIMENTS

In this section, we employed both synthetic and real-world data to evaluate the performance of the proposed algorithm. To evaluate the proposed algorithm, we compared it with three related state-of-the-art methods: NMF [22], $\ell_1$-NMF [48], and $\ell_{1/2}$-NMF [7]. It is worth to notice that all of the methods need to initialize the endmember matrix $\mathbf{X}$ and the abundance matrix $\mathbf{W}$ first. Here, we adopt VCA [17] to provide the initial endmember matrix $\mathbf{X}$ and then utilize the method in [49] to obtain the initial abundance matrix $\mathbf{W}$.

### A. Parameter Settings

Notice that the parameter $\sigma$ of the Gaussian kernel in Eq. (15) will affect the performance of the proposed method, and it is often determined empirically. In this study, we adopt the method in [38] to determine the parameter $\sigma$, which is computed by

$$\sigma^2 = \frac{\alpha}{2DN} \|\mathbf{Y} - \mathbf{XW}\|_F^2. \tag{29}$$

In this paper, the default value of $\alpha$ is taken as $\alpha = 1$.

There is another important parameter in our model: the regularization parameter $\lambda$ of the sparse penalty term. The value of parameter $\lambda$ is related to the sparseness of the material abundances.

However, the material abundances are the unknown variables. Here, we adopt the sparseness criteria in [11], [7], [50] to estimate $\lambda$ from hyperspectral data,

$$\lambda = \frac{1}{\sqrt{D}} \sum_{d=1}^{D} \frac{\sqrt{N} - \frac{\|\mathbf{y}^d\|_1}{\|\mathbf{y}^d\|_2}}{\sqrt{N} - 1}, \tag{30}$$

where $\mathbf{y}^d$ denotes the $d$-th band of the hyperspectral data.

### B. Evaluation Metrics

To evaluate the performance of the proposed method quantitatively, we adopt spectral angle distance (SAD) and root-mean-square error (RMSE) to measure the accuracy of the extracted endmembers and the corresponding abundances, respectively. These two metrics are widely adopted to evaluate the performance of the unmixing algorithms [5], [51], [7], [11]. The SAD is used to measure the distance between the ground true endmember and its estimated endmember, which is defined as follows:

$$\text{SAD}\left(\bar{\mathbf{x}}_k, \mathbf{x}_k\right) = \arccos\left(\frac{\bar{\mathbf{x}}_k^T \mathbf{x}_k}{\|\bar{\mathbf{x}}_k\|\|\mathbf{x}_k\|}\right), \tag{31}$$

where $\bar{\mathbf{x}}_k$ denotes the $k$-th true endmember and $\mathbf{x}_k$ is its estimated endmember ($\mathbf{x}_k$ is the $k$-th column of the endmember matrix $\mathbf{X}$).

Additionally, the RMSE is used to evaluate the distance between the ground true abundance map and its estimated abundance map, which is defined as follows:

$$\text{RMSE}\left(\bar{\mathbf{w}}^k, \mathbf{w}^k\right) = \left(\frac{1}{N}\|\bar{\mathbf{w}}^k - \mathbf{w}^k\|_2^2\right)^{\frac{1}{2}}, \tag{32}$$

where $\bar{\mathbf{w}}^k$ denotes the $k$-th true abundance map and $\mathbf{w}^k$ is its estimated abundance map ($\mathbf{w}^k$ is the $k$-th row of the abundance matrix $\mathbf{W}$).

### C. Experiments on Synthetic Data

In this subsection, we performed simulative experiments on synthetic data. The synthetic data is generated from a random selection of six true spectra from the United States Geological Survey (USGS) digital spectral library [52]. These six true signatures are shown in Fig. 1.

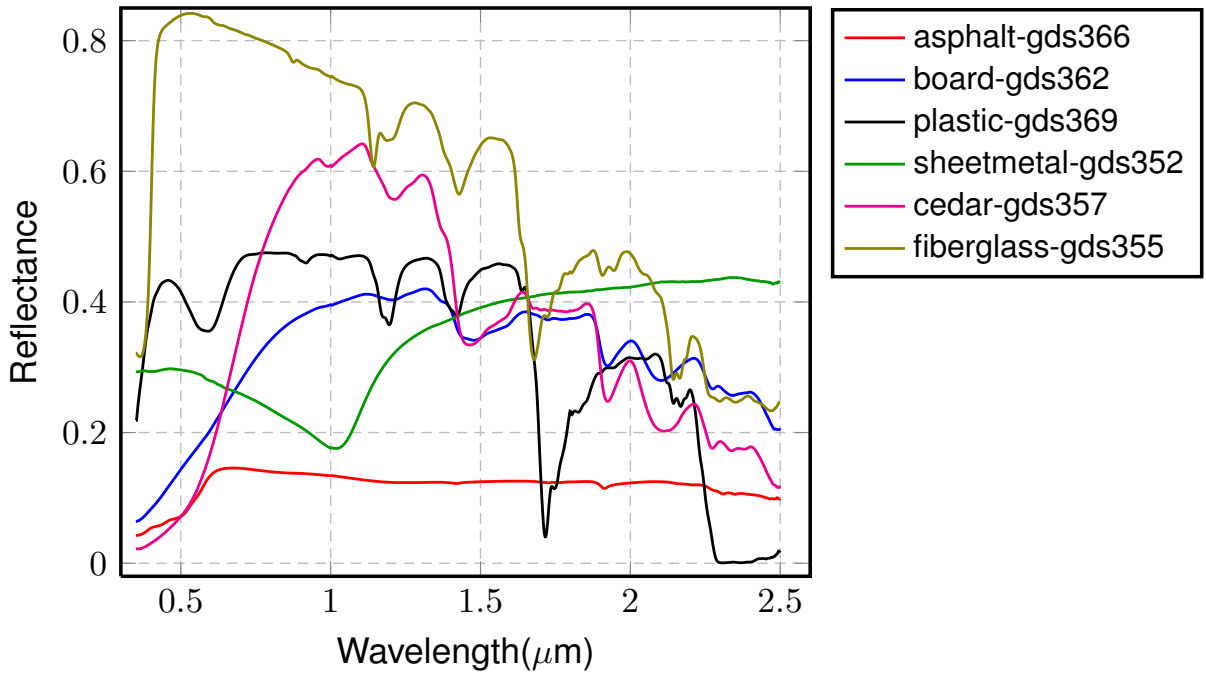

Figure 1. The selected endmember spectra used to generate synthetic data.

We adopt the method in [53] to generate the synthetic data. The generation process includes following steps:

1) An image of size $z^2 \times z^2$ is divided into units of $z \times z$ small blocks ($z = 8$ in this work).

2) Each block is filled up with the same type of ground cover, which is randomly selected from the six true spectra.

3) Utilize a $(z + 1) \times (z + 1)$ spatial low-pass filter to smooth the image, which generate mixed pixels.

4) Remove pure pixels and generate sparse abundances. Replace all the pixels whose largest abundance is larger than 0.8 with a mixture made up of only two endmembers with equal abundances.

5) Simulate noise. Add zero-mean Gaussian noise to the generate image data. The Signal-to-Noise Ratio (SNR) is defined as follows:

$$\text{SNR} = 10 \log_{10} \frac{\mathbb{E}\left[\mathbf{y}^T \mathbf{y}\right]}{\mathbb{E}\left[\mathbf{n}^T \mathbf{n}\right]}, \tag{33}$$

                                                                 

where $\mathbf{y}$ denotes the true spectra and $\mathbf{n}$ is the noise and $\mathbb{E}\left[\cdot\right]$ denotes the expectation operator. Here, we assume the noise is both spatially and spectrally uncorrelated. Given a particular value of SNR, the variance of the zero-mean Gaussian noise is

$$\sigma^2 = \frac{\mathbb{E}\left[\mathbf{y}^T\mathbf{y}\right]}{10^{\frac{\text{SNR}}{10}}}. \tag{34}$$

Notice that, to simulate the hyperspectral data in accordance with real situation, we add different levels of noise for different bands. Specifically, the SNR of each band is sampled from a Gaussian distribution $\text{SNR} \sim \mathcal{N}\left(\overline{\text{SNR}}, \epsilon^2\right)$. Here, we fix $\epsilon = 5$ in this work.

Here, we evaluate the performance of the proposed method in different noise levels. The proposed algorithm is compared with NMF, $\ell_1$-NMF and $\ell_{1/2}$-NMF. Figs. 2 and 3 show the performance results with different noise levels ($\overline{\text{SNR}} = \{50, 40, 30, 20, 10\}$). Fig. 2 presents the SAD curves of the four methods at different noise levels. Each sub-figure in Fig. 2 stands for the SAD of one of the endmembers. Fig. 3 presents the RMSE of the four methods at different noise levels. Each sub-figure in Fig. 3 denotes the RMSE of one of the abundance maps. It is obvious that our method ($\ell_1$-CENMF) achieves much lower values than other methods both in terms of SAD and RMSE. The main reason lies in that our method takes into account the difference of noise levels in different bands and models the noise in each band accurately. Therefore, our model performs significantly better than other methods. Meanwhile, as can be seen from Figs. 2 and 3, $\ell_1$-NMF and $\ell_{1/2}$-NMF outperform the original NMF method. This is mainly because the sparsity constraint is consistent with the sparseness of the abundance maps in the data.

### D. Experiments on Real Data

In this subsection, we applied our unmixing model to two real hyperspectral images: the HYDICE Urban data set and the Jasper Ridge data set. Both of the data sets contain noisy bands due to dense water vapor, atmospheric effects, and sensor noise.

Urban data set is one of the most widely used hyperspectral data for hyperspectral unmixing researches [5], [51], [7]. There are $307 \times 307$ pixels in this data set. It is composed of 210 spectral channels with a spectral resolution of 10nm from 400nm to 2500nm. As can be seen from Fig.

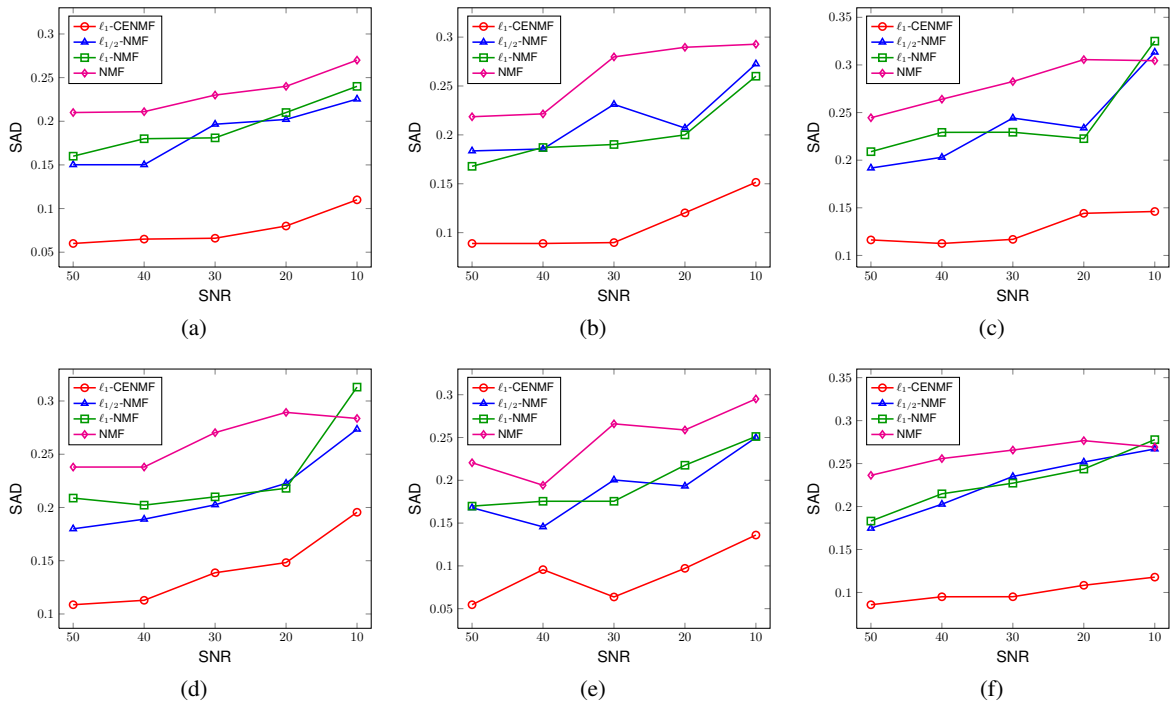

Figure 2. Comparison of the algorithms at different noise levels in terms of SAD. (a) asphalt-gds366. (b) board-gds362. (c) plastic-gds369. (d) sheetmetal-gds352. (e) cedar-gds357. (f) fiberglass-gds355.

4, the noise intensity of each band in this data set is different. Meanwhile, we can obviously observe that the SNR of each band in this figure satisfies $SNR_{205} > SNR_{206} > SNR_{207} > SNR_{208}$. Actually, the bands $1-4$, $76$, $87$, $101-111$, $136-153$ and $198-210$ are corrupted by serious noise. For most of the existing unmixing methods, these bands with low SNR should be removed before performing unmixing algorithms. Here, we take all of bands to perform the unmixing algorithms.

According to the ground truths [54] of the data, there are four interesting endmembers: Tree, Roof, Asphalt and Grass. Fig. 5 presents the intuitive compassion of the abundance maps obtained by all of the methods. As can be seen from this figure, the abundance maps obtained by NMF, $\ell_1$-NMF and $\ell_{1/2}$-NMF contain more noise than the results of our method. The reason is that NMF, $\ell_1$-NMF and $\ell_{1/2}$-NMF consider that all bands in the data contribute equally to the unmixing procedure, thus the noisy bands will significantly affect the unmixing results. By comparison, our model, automatically and adaptively assign small weights to those bands with low SNR, is more insensitive to noisy bands. As can be seen from Fig. 5, our model obtains the most similar abundance maps compared with the ground truths. The quantitative comparison of SAD[2] and RMSE on the Urban data is shown in Table I. From this table, we see that our $\ell_1$-CENMF model obtains much better results than other methods.

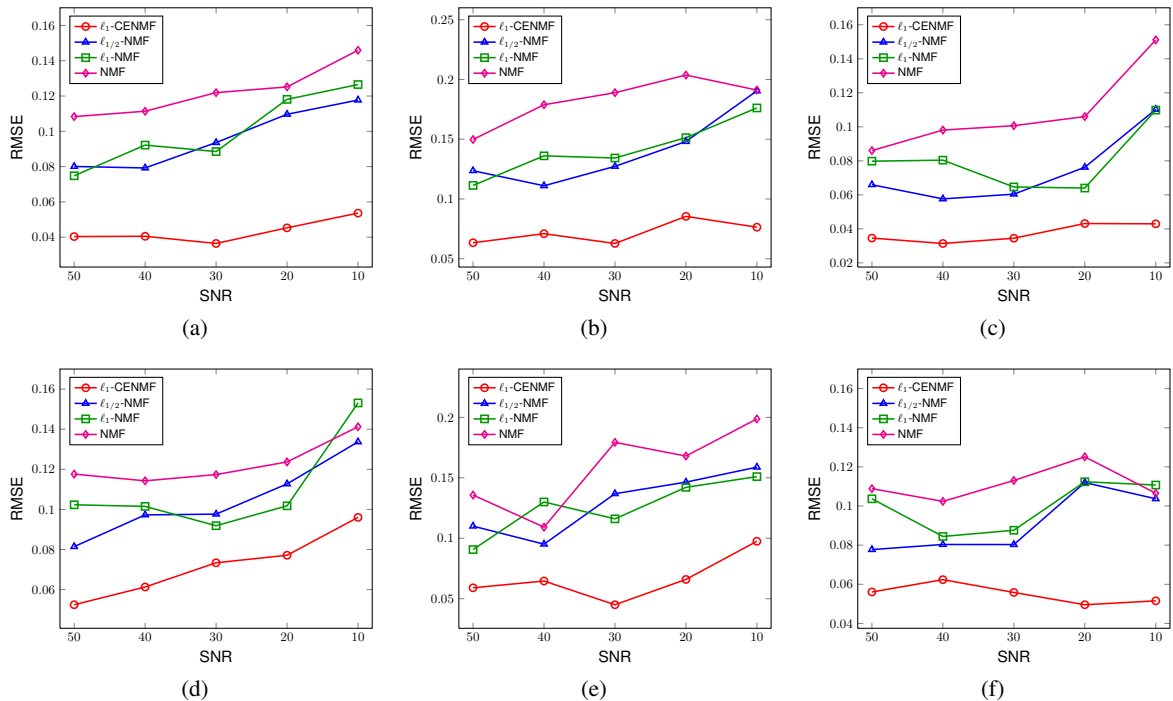

Figure 3. Comparison of the algorithms at different noise levels in terms of RMSE. (a) asphalt-gds366. (b) board-gds362. (c) plastic-gds369. (d) sheetmetal-gds352. (e) cedar-gds357. (f) fiberglass-gds355.

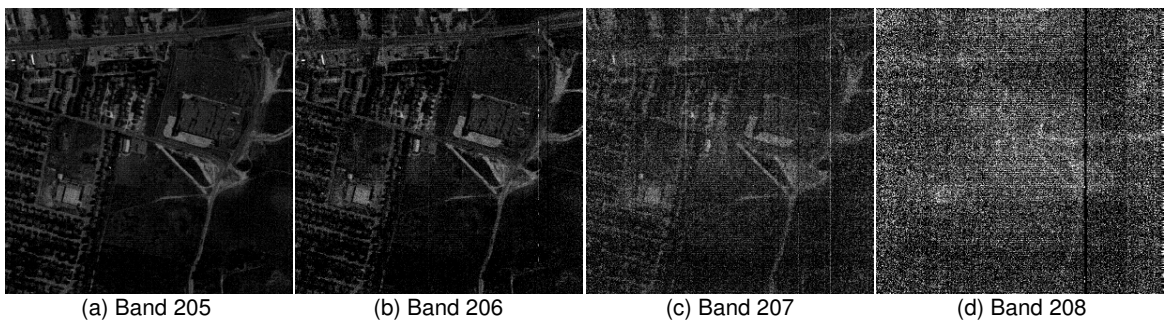

|          |          |          |          |
|----------|----------|----------|----------|
| (a) Band 205 | (b) Band 206 | (c) Band 207 | (d) Band 208 |

Figure 4. Some noisy bands in Urban data set. The noise level in this data set differs from band to band.

---

[2]Here, we removed the spectra of the noise corrupted bands (bands 1-4, 76, 87, 101-111,136-153 and 198) before computing the SAD.

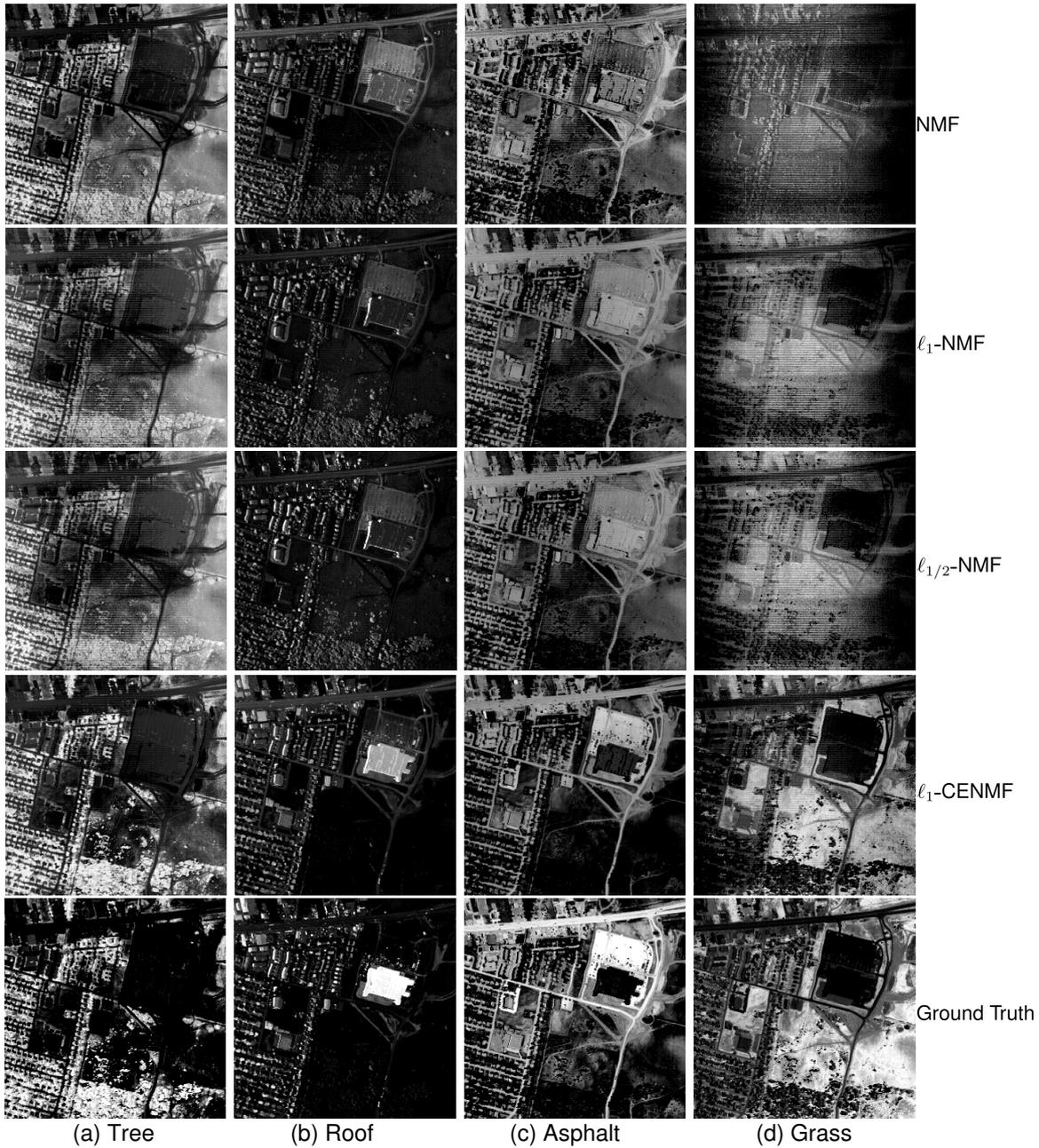

Figure 5. Abundance maps on the Urban data set. From the first row to the fourth row denote the abundance maps obtained by NMF, $\ell_1$-NMF, $\ell_{1/2}$-NMF and our $\ell_1$-CENMF, respectively. The last row illustrates the Ground Truths.

After the optimization algorithm of our model converged, we can get the weight of each band according to the auxiliary vector $\mathbf{u}$. Fig. 6 plots the weights of all bands in Urban data set. It is clearly seen that the bands with small weights correspond to the bands with serious noise

Table I
The SAD and RMSE of the four methods on Urban data set.

| Endmember | SAD | | | | RMSE | | | |
|---|---|---|---|---|---|---|---|---|
| | NMF | $\ell_1$-NMF | $\ell_{1/2}$-NMF | $\ell_1$-CENMF | NMF | $\ell_1$-NMF | $\ell_{1/2}$-NMF | $\ell_1$-CENMF |
| Tree | 0.1654 | 0.1403 | 0.0923 | **0.0882** | 0.2930 | 0.2924 | 0.2572 | **0.1496** |
| Roof | 0.5609 | 0.4924 | 0.3019 | **0.2576** | 0.2017 | 0.2014 | 0.1914 | **0.1137** |
| Asphalt | 0.3291 | 0.3165 | 0.2570 | **0.1090** | 0.2639 | 0.2242 | 0.2239 | **0.1590** |
| Grass | 0.4509 | 0.3861 | 0.3627 | **0.2209** | 0.4015 | 0.3557 | 0.3546 | **0.1928** |
| Mean | 0.3766 | 0.3338 | 0.2535 | **0.1689** | 0.2901 | 0.2684 | 0.2568 | **0.1538** |

(bands $1 - 4$, $76$, $87$, $101 - 111$, $136 - 153$ and $198 - 210$). Here, we show the image of the $105$-th band of Urban data set in Fig. 7. As can be seen from this figure, this band is corrupted by serious noise. Meanwhile, this band is assigned a small weight as illustrated in Fig. 6. This demonstrates that our correntropy based model can effectively cope with the noise bands in the hyperspectral image.

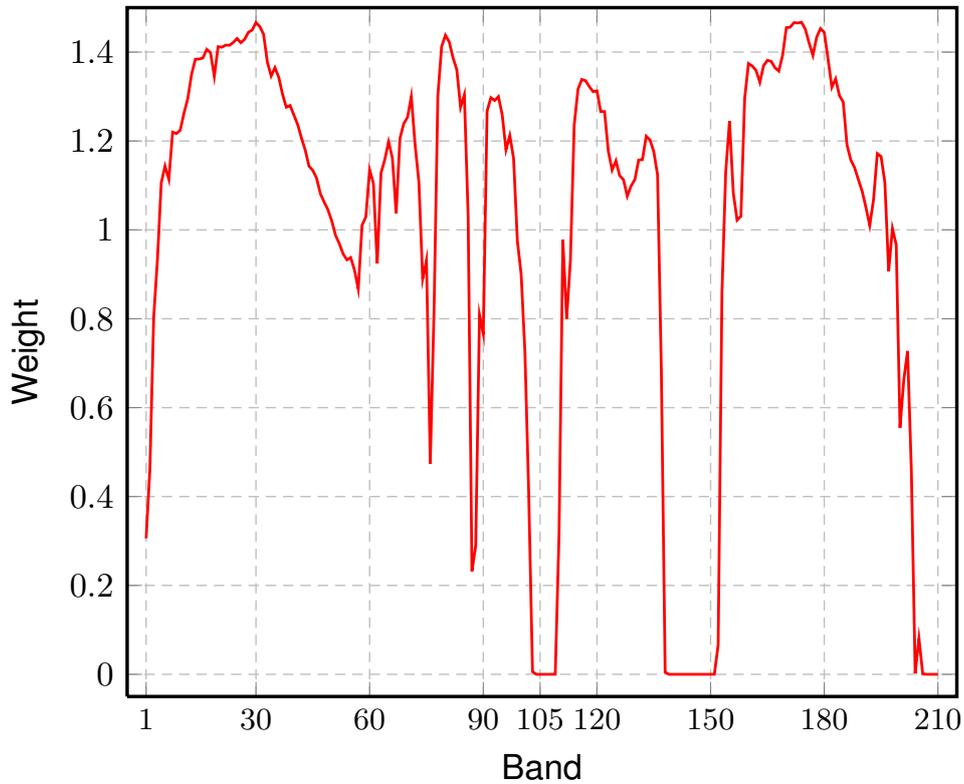

Figure 6. The weights of all bands in Urban data.

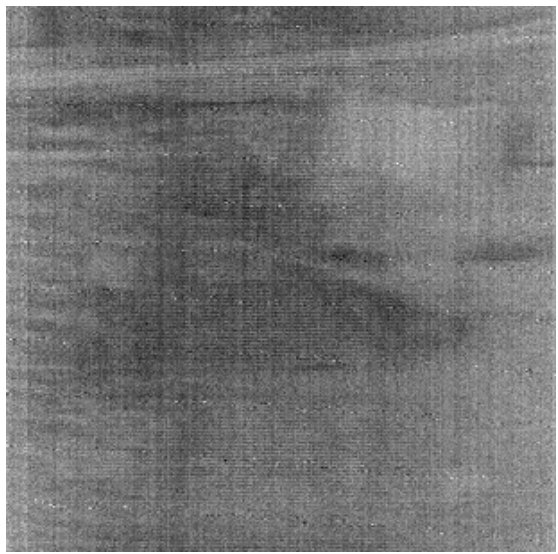

Figure 7. The image of the105-th band in Urban data set.

The other real data is the Jasper Ridge data set. It is another popular hyperspectral data set introduced in ENVI Tutorials [55] for hyperspectral analysis. It is composed of 224 spectral channels covering the wavelength from 380nm to 2500nm, with a spectral resolution of 10nm. After the black bands (bands 1, 108-112 and 154-166) are removed, only 205 bands remain. The selected subregion from the Jasper Ridge data set is shown in Fig. 8, which contains $100 \times 100$ pixels. There are four interesting endmembers: Water, Tree, Soil and Road. It can be observed from Fig. 8 that this data set contains noisy bands and the noise level in each band is different from those of the others.

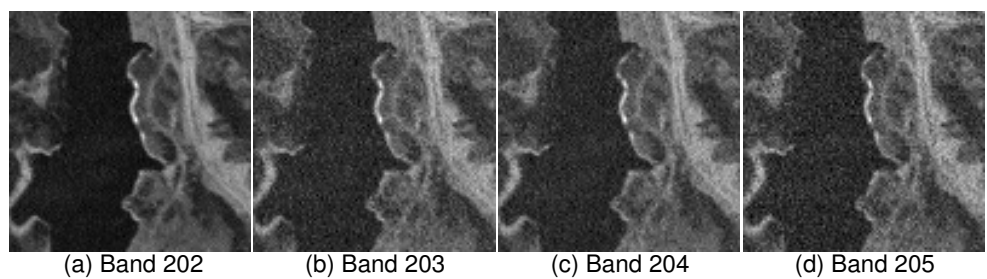

(a) Band 202    (b) Band 203    (c) Band 204    (d) Band 205

Figure 8. Some bands in Jasper data set.

Fig. 9 presents the abundance maps obtained by all the methods. This figure shows that our $\ell_1$-CENMF method yields the most similar abundance maps compared with the ground truths.

Table II shows the quantitative comparison in terms of SAD and RMSE on the Jasper Ridge

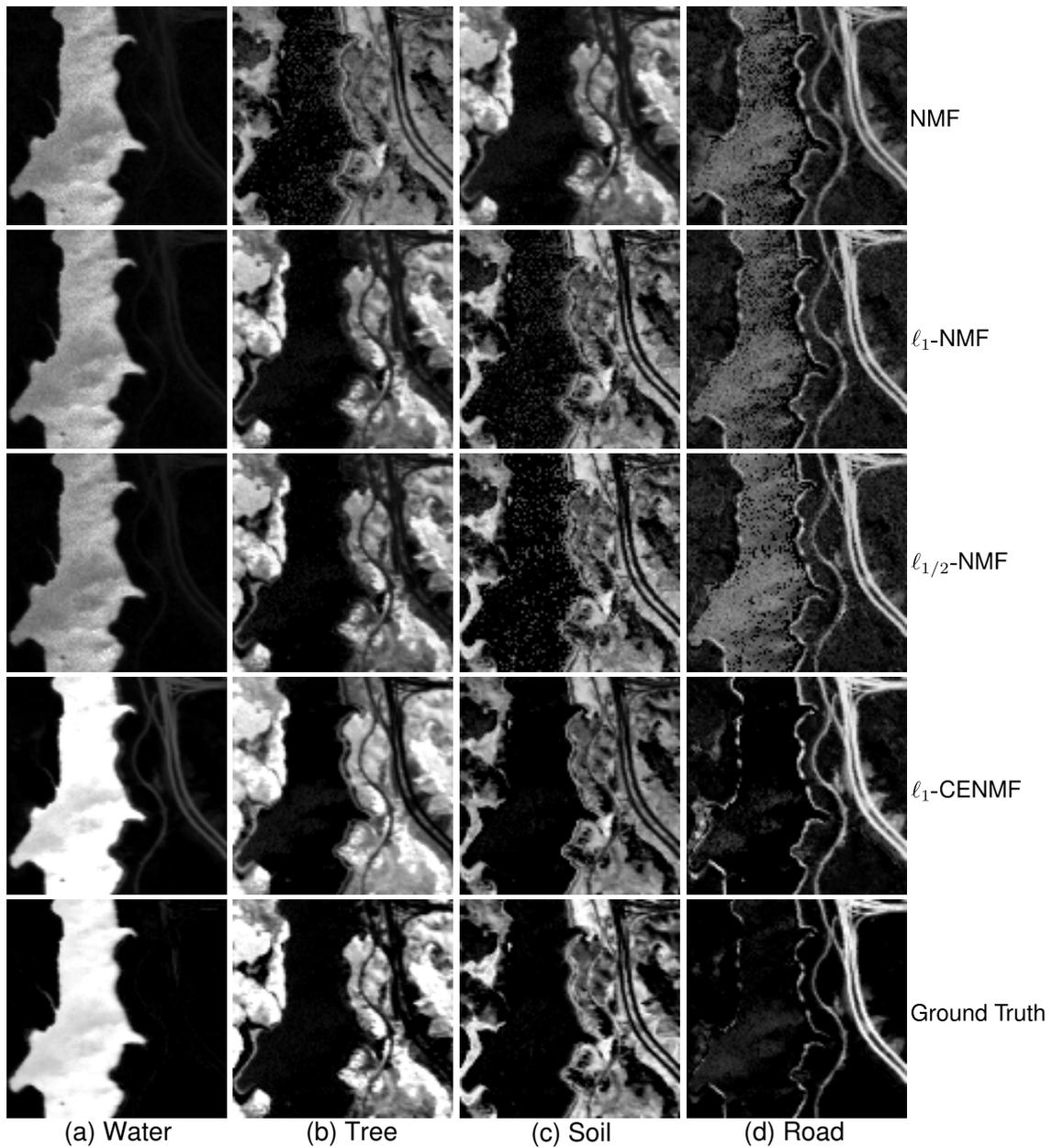

(a) Water       (b) Tree       (c) Soil       (d) Road

Figure 9. Abundance maps on Jasper Ridge data set. From the first row to the fourth row denote the abundance maps obtained by NMF, $\ell_1$-NMF, $\ell_{1/2}$-NMF and our $\ell_1$-CENMF, respectively. The last row illustrates the Ground Truths.

data. It is shown that our $\ell_1$-NMF obtains the best performance for most of the endmembers and their corresponding abundance maps.

Table II
THE SAD AND RMSE OF THE FOUR METHODS ON JASPER DATA SET.

| Endmember | SAD | | | | RMSE | | | |
|---|---|---|---|---|---|---|---|---|
| | NMF | $\ell_1$-NMF | $\ell_{1/2}$-NMF | $\ell_1$-CENMF | NMF | $\ell_1$-NMF | $\ell_{1/2}$-NMF | $\ell_1$-CENMF |
| Water | 0.1081 | 0.0935 | 0.0915 | **0.0602** | 0.1791 | 0.1764 | 0.1958 | **0.0707** |
| Tree | 0.1336 | 0.1213 | 0.1285 | **0.1155** | 0.1214 | **0.1041** | 0.1088 | 0.1121 |
| Soil | 0.2194 | 0.2110 | 0.1713 | **0.1472** | 0.1312 | 0.1018 | 0.1034 | **0.0904** |
| Road | 0.2046 | 0.1713 | 0.1618 | **0.1015** | 0.1837 | 0.1782 | 0.2092 | **0.0949** |
| Mean | 0.1643 | 0.1493 | 0.1382 | **0.1061** | 0.1539 | 0.1401 | 0.1543 | **0.0920** |

We also present the weights of all bands in Jasper Ridge data in Fig. 10. It can be observed that the first band is assigned a small weight. We show the first band of the Jasper Ridge data in Fig. 11. As can be seen from this figure, this band contains high-level noise. This illustrates that our model can adaptively assign small weights to the bands with serious noise, which is the main reason that our $\ell_1$-CENMF outperforms the other methods.

In order to further demonstrate the effectiveness of the proposed model, we perform another experiment on Urban data set. In this experiment, we first remove the noisy bands (bands $1-4$, 76, 87, $101-111$, $136-153$ and $198-210$), thus only 162 bands remain. Then, $\ell_{1/2}$-NMF and our $\ell_1$-CENMF are applied to unmix this data. Table III reports the performances of the two methods in terms of SAD and RMSE. For comparison, Table III also shows the quantitative results of Urban data set with all of the bands have been used to perform spectral unmixing. Here, the table heading (162) denotes the Urban data with 162 bands, and the table heading (all) denotes the Urban data with all bands. From the results listed in Table III, we can conclude the following three points. First, $\ell_{1/2}$-NMF achieves better results on Urban data with 162 bands than that with all bands. The reason is that $\ell_{1/2}$-NMF considers that all bands contribute equally to spectral unmixing, thus the noisy bands will significantly affect the unmixing results. This implies that $\ell_{1/2}$-NMF is very sensitive to noisy bands. Second, the results of our $\ell_1$-CENMF on the Urban data with 162 bands and all bands are very close. The reason is that, the noisy bands, which are adaptively assigned to small weights, have little effect to the objective function. Accordingly, our method is very robust to noisy bands. Third, as demonstrated in Table III, our method has better performance both on the data of 162 bands and all bands than $\ell_1$-NMF does on the data of 162 bands. The reasons are twofold. On the one hand, the removed bands are manually selected, which leads to discarding available spectral information. As illustrated in Fig. 4, although there is some noise in the 205-th band, it contains useful spectral information. Our method can effectively and adaptively exploit the useful spectral information from all of the bands. On the other hand, due to the fact that the noise level in different bands is different and manually removing noisy bands is primarily dependent on subjective observation, the rest of the bands may still contains noise.

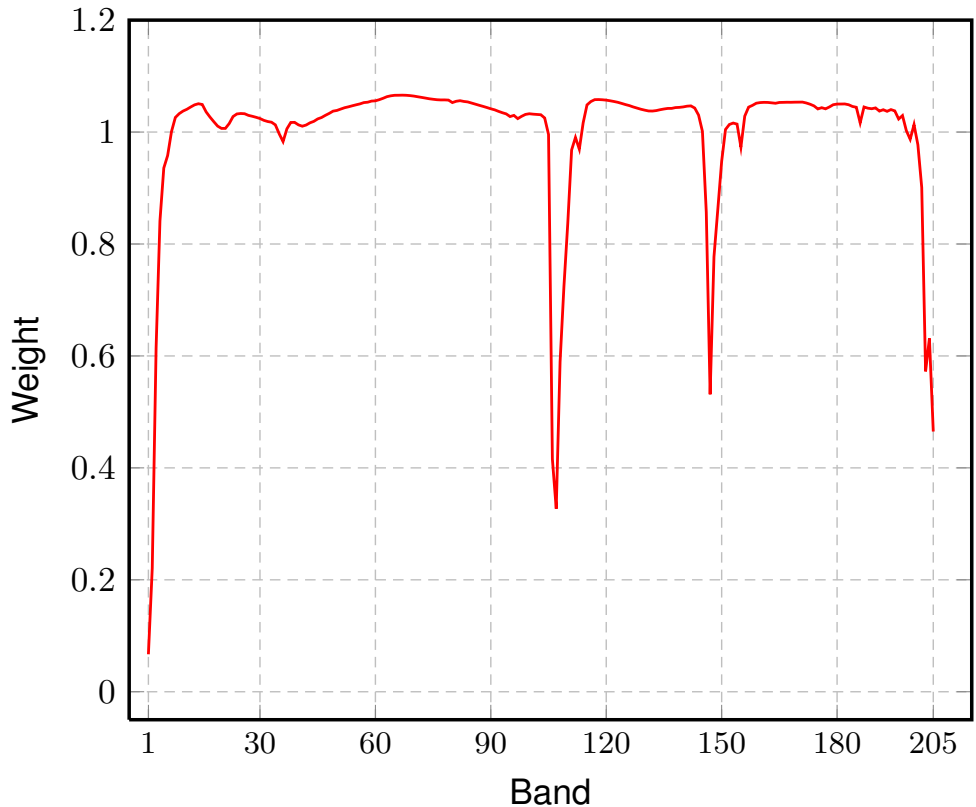

Figure 10. The weights of all bands in Jasper Ridge data.

In summary, our method can adaptively assign small weights to bands with low SNR and give more emphasis on bands with high SNR. The adaptive capacity of our model can utilize the data more effectively for hyperspectral unmixing, and thus facilitates its real-world applications.

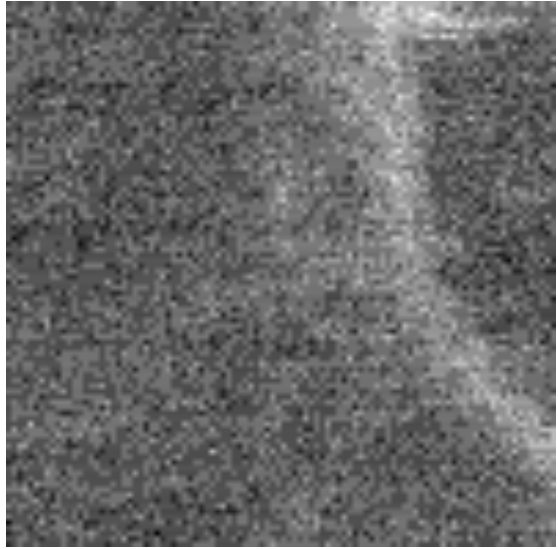

Figure 11. The image of the first band in Jasper Ridge data.

Table III
THE SAD AND RMSE OF $\ell_{1/2}$-NMF AND $\ell_1$-CENMF METHODS ON URBAN DATA SET. THE COLUMN HEADING (162) DENOTES THE URBAN DATA WITH 162 BANDS, AND THE COLUMN HEADING (ALL) DENOTES THE URBAN DATA WITH ALL BANDS.

| Endmember | SAD | | | | RMSE | | | |
|---|---|---|---|---|---|---|---|---|
| | $\ell_{1/2}$-NMF (all) | $\ell_{1/2}$-NMF (162) | $\ell_1$-CENMF (all) | $\ell_1$-CENMF (162) | $\ell_{1/2}$-NMF (all) | $\ell_{1/2}$-NMF (162) | $\ell_1$-CENMF (all) | $\ell_1$-CENMF (162) |
| Tree | 0.0923 | 0.0751 | 0.0882 | **0.0699** | 0.2572 | 0.1488 | 0.1496 | **0.1407** |
| Roof | 0.3019 | 0.2513 | 0.2576 | **0.2457** | 0.1914 | 0.1350 | **0.1137** | 0.1250 |
| Asphalt | 0.2570 | 0.1409 | **0.1090** | 0.1258 | 0.2239 | 0.1586 | 0.1590 | **0.1391** |
| Grass | 0.3627 | 0.2285 | 0.2209 | **0.2143** | 0.3546 | 0.2296 | **0.1928** | 0.2028 |
| Mean | 0.2535 | 0.1739 | 0.1689 | **0.1640** | 0.2568 | 0.1581 | 0.1538 | **0.1519** |

## V. CONCLUSIONS

In this paper, we have proposed a robust hyperspectral unmixing algorithm based on correntropy based metric. The main contributions of this work were as follows: (1) the proposed model utilizes correntropy based metric to define the objective function, which can adaptively assign small weights to noisy bands and give more emphasis on noise-free bands. Thus, our model is much more insensitive to noise; (2) a convergent optimization algorithm is proposed based on half-quadratic optimization technique. Additionally, theoretical analyses about the derived model for its robustness were given. Comparative experiments on both synthetic and real-world data show that the proposed model achieves much better results than the state-of-the-art methods.

Although the optimization algorithm of the proposed method is convergent, the convergence is slow. The main reason is that the multiplicative update algorithm was adopted to solve the subproblem (16) in step 2 of the alternate optimization procedure, which suffers from drawback of slow convergence. In the future, we would like to speed up our algorithm by adopting more efficient optimization techniques.